\documentclass{article} 
\usepackage{iclr2020_conference,times}

\usepackage{amsmath,amsfonts,bm}

\def\eqref#1{equation~\ref{#1}}

\def\1{\bm{1}}

\def\vx{{\bm{x}}}

\DeclareMathAlphabet{\mathsfit}{\encodingdefault}{\sfdefault}{m}{sl}
\SetMathAlphabet{\mathsfit}{bold}{\encodingdefault}{\sfdefault}{bx}{n}

\usepackage{graphicx}
\usepackage{comment}
\usepackage{amsmath,amssymb} 
\usepackage{hyperref}
\usepackage{url}
\usepackage{soul}
\usepackage{graphicx}
\usepackage{amsthm}
\usepackage{booktabs}
\usepackage{bbold}
\usepackage{enumitem}
\usepackage{epsfig}
\usepackage{bm}
\usepackage{amstext}
\usepackage{booktabs}
\usepackage[ruled,vlined]{algorithm2e}
\SetKwInput{kwInit}{Init}
\SetKwInput{kwRequire}{Require}
\SetKwComment{Comment}{$\triangleright$\ }{}

\SetCommentSty{mycommfont}
\newcommand{\nonl}{\renewcommand{\nl}{\let\nl\oldnl}}
\usepackage{amsfonts}
\usepackage{multirow}
\usepackage{subcaption}
\usepackage{mathtools}
\usepackage{enumerate}

\newcommand{\mc}{\mathcal}

\newcommand{\Dtrain}[1]{D^{tr}_{#1}}

\usepackage{array}
\newcolumntype{P}[1]{>{\centering\arraybackslash}p{#1}}
\def\1{\bm{1}}
\def\vx{{\bm{x}}}

\title{Bilevel Continual Learning}

\author{Quang Pham$^1$, Doyen Sahoo$^2$, Chenghao Liu$^1$, Steven C.H. Hoi $^{1,2}$ \\
$^1$ Singapore Management University \\
\texttt{\{hqpham.2017,chhoi\}@smu.edu.sg twinsken@gmail.com} \\
$^2$ Salesforce Research Asia\\
\texttt{dsahoo@salesforce.com}
}

\iclrfinalcopy 
\begin{document}

\maketitle

\begin{abstract}
Continual learning aims to learn continuously from a stream of tasks and data in an online-learning fashion, being capable of exploiting what was learned previously to improve current and future tasks while still being able to perform well on the previous tasks. One common limitation of many existing continual learning methods is that they often train a model directly on all available training data without validation due to the nature of continual learning, thus suffering poor generalization at test time. In this work, we present a novel framework of continual learning named ``Bilevel Continual Learning" (BCL) by unifying a {\it bilevel optimization} objective and a {\it dual memory management} strategy comprising both episodic memory and generalization memory to achieve effective knowledge transfer to future tasks and alleviate catastrophic forgetting on old tasks simultaneously. 
Our extensive experiments on continual learning benchmarks demonstrate the efficacy of the proposed BCL compared to many state-of-the-art methods. Our implementation is available at \href{https://github.com/phquang/bilevel-continual-learning}{https://github.com/phquang/bilevel-continual-learning}.

\end{abstract}

\section{Introduction}
Unlike humans, conventional machine learning methods, particularly neural networks, struggle to learn continuously because these models lose their abilities to perform acquired skills when they learn a new task \citep{french1999catastrophic}. Continual learning systems are specifically designed to learn continuously from a stream of tasks. They are able to accumulate knowledge over time to improve the future learning outcome, while still being able to perform well on the previous tasks. In the literature, prior works mainly focus on the continual learning protocol where the whole task data arrives at each step and the learner is allowed to train the current task on many epochs.
This does not well reflect the real-world scenarios where data arrives sequentially and the learner has to learn new tasks on the fly.
In this work, we make a next step towards the more realistic continual learning by developing our methods in the {\it online continual learning} regime where the training of each task is also performed in an online fashion with data arrives sequentially \citep{lopez2017gradient}.
Such a protocol is more appealing as optimizing neural networks usually requires a lot of training episodes and various techniques such as data augmentations, learning rate scheduling, etc. while struggling when data arrives in an online fashion \citep{sahoo2018online}.

To be able to learn in online continual learning, the model not only has to prevent catastrophic forgetting but also leverage its past knowledge to improve the learning of the current task. It is important to balance both aspects so that the performance on all tasks is maximized. Despite the initial success of existing works \citep{chaudhry2019agem,chaudhry2019tiny,hou2018lifelong,lopez2017gradient,riemer2018learning}, there is still a huge performance gap because they struggle to balance between knowledge transfer and preventing catastrophic forgetting. Existing methods either favor improving knowledge transfer \citep{chaudhry2019tiny,riemer2018learning} or focus on preventing catastrophic forgetting \citep{hou2018lifelong}.
Moreover, given that we cannot store all information of old tasks, the model may not generalize well at test time because of the information loss from the limited memory. Therefore, it is important to balance between alleviating catastrophic forgetting and facilitating knowledge transfer, especially in the online setting.

To address the aforementioned challenges, we based on the cross-validation principle \citep{jenni2018deep} and propose {\it Bilevel Continual Learning} (BCL), which formulates continual learning as a problem of improving the model's generalization on a separate set of data from all observed tasks. During training, BCL maintains two disjoint memory units: an episodic memory for training and a generalization memory for improving generalization. Importantly, the generalization memory is never used to directly train the main model but only for improving its generalization. BCL learns new samples by first initializes a fast-weight and train it with experience replay using the episodic memory. Then, the trained fast-weight is used to update the original model such that it can generalize to the generalization memory. Therefore, BCL uses a {\it bilevel optimization} objective \citep{colson2007overview} with the inner problem as experience replay with the current data and the {\it outer problem} as optimizing the model's performance on the generalization memory. As a result, BCL alleviates catastrophic forgetting because it is optimized to generalize to previous tasks. Similarly, it facilitates knowledge transfer because the loss on the current task's unseen data is minimized. We develop a practical first-order approximation of the bilevel continual learning problem that can apply on large, deep neural networks. Furthermore, we address the bias in the inner optimization problem caused by the small episodic memory size. Consequently, our BCL algorithms strike a great balance between alleviating catastrophic forgetting and facilitating the learning of future tasks. 

Interestingly, our BCL design is also related to the Complementary Learning Systems (CLS) \citep{mcclelland1995there,kumaran2012generalization}, which is an important approach to continual learning. Particularly, our dual memory design corresponds to the {\it episodic memory} and the {\it semantic memory} in the brain. Moreover, the fast-weight to learn new samples plays the {\it hippocampus} role in rapid learning and acquiring new experiences. The learned knowledge from the trained fast-weight is then consolidated to the base model such that it can generalize to the generalization (semantic) memory, which is never revealed to the fast-weights. Under BCL training, the main model's role resembles the {\it neocortex} of capturing the common knowledge of all observed tasks.

In summary, our contributions are as follows. First, we propose a novel continual learning objective based on bilevel optimization and a dual memory management strategy. Second, we derive a practical algorithm based on a first-order approximation, which can be applied to large models efficiently. Finally, we conduct comprehensive experiments on several online continual learning benchmarks to validate the efficacy of our proposed algorithm against a suite of continual learning baselines.

\begin{figure}[t]
	\vskip 0.1in
	\begin{center}
		\centerline{\includegraphics[width=0.90\textwidth]{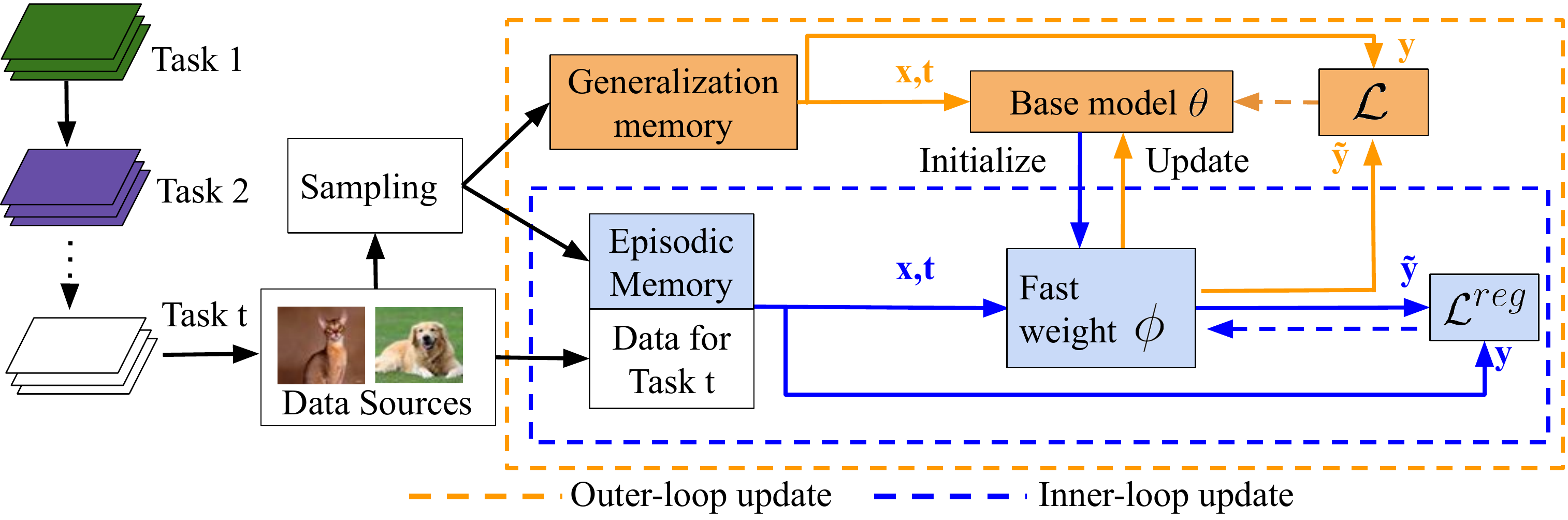}}
		\caption{Bilevel Continual Learning optimizes the main model  $\bm \theta$ to perform well on the generalization memory using the fast weight $\bm \phi$ and the episodic memory.}
		\label{fig:bcl}
	\end{center}
	\vskip -0.2in
\end{figure}

\section{Bilevel Continual Learning}
\label{sec:mkd}
In this work, we propose Bilevel Continual Learning (BCL), a conceptually new framework for continual learning based on bilevel optimization \citep{colson2007overview}. Different from existing works that optimize a model to perform well on the training data, we directly aim at improving the model's generalization across tasks, given that only the training data are observed. BCL maintains two disjoint memory units: an episodic memory $\mc M^{em} = \cup \mc M^{em}_t$ for experience replay and a generalization memory $\mc M^{gm}$ that is never used to directly trained the model but only for improving the generalization. When a new training sample arrives, BCL initializes a fast weight $\bm \phi$ from the main model $\bm \theta$ to learn this sample through experience replay with the episodic memory. Then, the trained fast weight is used to update the main model $\bm \theta$ such that is can perform well on the generalization set. Fig.~\ref{fig:bcl} illustrates our proposed BCL framework.

\subsection{Bilevel Learning}
For each incoming mini batch of data $\mc B^t_n$ of task $\mc T_t$, BCL initializes a fast weight $\bm \phi$ to acquire the new knowledge in $\mc B^t_n$. The trained fast weight $\bm \phi^*$ is then used to update $\bm \theta$ such that it can generalize to the generalization memory. This training objective can be formulated as the following bilevel optimization:
\begin{align}
\min_{\bm \theta} \;\; & \mc L^{outer}(\bm \phi^* (\bm \theta), \mc M^{gm} ; \bm \theta),  \notag \\
\mathrm{ s.t. } \;\; & \bm \phi^* = \arg\min_{\bm \phi}  \mc L^{inner}(\bm \phi, \mc B_n^{t} \cup \mc M^{er}_{n}; \bm \theta), 
\label{eqn:meta-obj}
\end{align}
where $\mc B^t_n$ is the $n-th$ data of the current task $\mc T_t$.
Each $\bm \theta$ will parameterize an inner optimization problem $\mc L^{train}_{\bm \theta}(\cdot)$ which we optimize with respect to $\bm \phi$. Once we obtained a solution $\bm \phi^*$ of the inner problem, we then optimize the outer problem with respect to $\bm \theta$. In this work, we use the cross-entropy loss for both the inner and outer problems: 
\begin{equation}
\mc L(\bm \phi, (\vx,y,t);\bm \theta) =  \mathrm{KL}(y || f_{\phi}(\vx,t)) + \mathrm{const},
\label{eqn:loss}
\end{equation}
where the right-hand side of Eq.~(\ref{eqn:loss}) depends on $\bm \theta$ as we initialize $\bm \phi$ from $\bm \theta$.
The cross-validation principle and the bilevel optimization formulation in Eq.(~\ref{eqn:meta-obj}) also appears in different problems such as supervised learning \citep{jenni2018deep}, hyper-parameter optimization \citep{franceschi2017forward}, and architecture search \citep{liu2018darts}. 

\vspace*{-0.1in}
\subsection{First-Order Approximation}
\vspace*{-0.1in}
In general, solving bilevel problems such as Eq.~(\ref{eqn:meta-obj}) is challenging for neural networks due to the exact solution of the inner optimization. 
While existing methods \citep{domke2012generic,jenni2018deep} was developed to train neural networks on a single task, they might not be feasible solutions when tasks arrive sequentially. Another compelling approach is approximating $\bm \phi^*$ by training $\bm \phi$ using only a few gradient steps \citep{liu2018darts}. This approximation is more suitable in our setting since data arrives in small batches, and a few gradient steps can achieve a reasonably good performance. Therefore, we decide to adopt and further develop this approximation in our work.
Particularly, the inner problem for each incoming training sample $\mc B^t_n$ is solved by: 
\begin{align}
\bm \phi_{i} \gets \bm \phi_i -\alpha \nabla_{\bm \phi_i}\mc L(\bm \phi_i, \mc B^t_n \cup \mc M^{em}_n; \bm \theta) \; \mathrm{ where } \; \bm \phi_0 \gets \bm \theta.
\label{eqn:maml-inner}
\end{align}
After obtaining $\bm \phi^*$ by Eq.(~\ref{eqn:maml-inner}), the outer optimization for $\bm \theta$ can be obtained by the chain rule: 
\begin{align}\label{eqn:maml-outer}
\bm \theta_t \gets & \bm \theta_t - \beta  \nabla_{\bm \theta} \mc L(\bm \phi^*, \mc M^{gm})  \notag \\
\gets & \bm \theta_t - \beta \frac{\partial \bm \phi^*}{\partial \bm \theta} \cdot \frac{\partial} {\partial {\bm \phi^*}} \mc L(\bm \phi^*, \mc M^{gm})
\end{align}
Unfortunately, the expression in Eq.(~\ref{eqn:maml-outer}) is very expensive in practice due to the Hessian-vector product in the second term. To alleviate the computational cost, we use the first-order approximation proposed in \citep{nichol2018first,zhang2019lookahead}. Particularly, the outer optimization is obtained by interpolating only in the parameter space:
\begin{align} \label{eqn:first-order}
	\bm \theta_t = & \bm \theta_t + \beta (\bm \phi' - \bm \theta_t), \notag \\
	\mathrm{where} \; \bm \phi' = & \bm \phi^* - \alpha \nabla_{\bm \phi^*}\mc L(\bm \phi^*, \mc M^{gm})
\end{align} 
In Eq.(~\ref{eqn:first-order}), we first obtain a one-step look-ahead parameter $\bm \phi'$ from $\bm \phi^*$ and then update $\bm \theta$ by linearly interpolate between the current $\bm \theta$ and $\bm \phi'$.
It is common in practice to perform several SGD steps in Eq.~(\ref{eqn:maml-inner}) and to obtain a good quality fast weight $\bm \phi_i$ Moreover, we always keep the main model $\bm \theta$ while $\bm \phi$ is created and then discarded after each outer update.

\subsection{Preventing The Inner Optimization Bias}
During the inner optimization, current task data is mixed with previous data in the episodic memory for experience replay training. However, previous data in the episodic memory are limited, which creates a bias towards the current task, which has more training data.
Such bias will drive the model towards the current task, resulting in a performance degrade.

To reduce this bias, we propose to regularize the inner optimization by preventing the fast-weight $\bm \phi$ from deviating too much from the previous main models $\{\bm \theta_{<t}\}$ in the predictive distribution space. This can be achieved by employing a knowledge distillation regularizer \citep{hinton2015distilling} on the inner objective as follows:
\begin{align}\label{eqn:FTRML-cls}
\mc L^{reg}_{\bm \theta} (\bm \phi_i, \mc B^t_n \cup \mc M^{er}_n) \leftarrow \mc L_{\bm \theta} (\bm \phi_i, \mc B^t_n \cup \mc M^{er}_n) + \sum_{\mathclap{\vx,y \in \mc M^{er}_n}} \mathrm{KL}_{\tau}(p_{\bm \theta_j}(y | \vx) || p_{\bm \phi_i}(y | \vx)) ,
\end{align}
where $\tau$ denotes the softmax's temperature, which is usually set to be greater than 1.
Eq.~(\ref{eqn:FTRML-cls}) shows that we can achieve our regularization goal by minimizing the empirical KL divergence between the two predictive distributions between the main and the fast-weight models. Calculating the regularizer in Eq.~(\ref{eqn:FTRML-cls}) requires the memory to store the triplets $(\vx,y,p_{\bm \theta_i}(y|\vx))$. However, this incurs insignificant computational and memory costs since no additional forward/backward pass is performed and the dimension of $p_{\bm \theta_j}(y|\vx)$ is much smaller than that of $\vx$. 

\subsection{Bilevel Continual Learning Algorithm}
\begin{figure}[!t]
	\begin{algorithm}[H]
		\DontPrintSemicolon
		\kwInit{$\bm \theta_1$, $\mc M \gets \varnothing$}
		\kwRequire{Memory management strategy for $\mc M$}
		\For{$t \gets 1$ \textbf{to} $T$ }{
			{Observe the dataset $\mc D^{tr}_t$ sequentially}\;
			\For{$n \gets 1$ \textbf{to} $n_{\text{batches}}$}{	
				{Receive a mini batch of data $\mc B_n$ from $\mc D^{tr}_t$ }\;
				{Randomly sample $\vx^{eval}, y^{eval}$ from $\mc B_n$}\;
				{$\mc M^{gm} \leftarrow $ update $\mc M^{gm} \text{ with } \{\vx^{eval},y^{eval} \} $}\;
				{$\mc B_n \leftarrow \mc B_n \setminus \{\vx^{eval}, y^{eval} \} $}\;
				\For(\Comment*[f]{outer loop}){$j \gets 1$ \textbf{to} $n_{\mathrm{outer}}$ }  {	
					{$\bm \phi_{0} \leftarrow \bm \theta_t$} \Comment*[r]{Initialize the fast-weight}
					{$\mc B_{\mc M} \leftarrow Sample(\mc M)$}\;
					{$\bar{\mc M}_j =  B_{\mc M} \cup \mc B_n$}\;
					\For(\Comment*[f]{inner loop}){$i \gets 1$ \textbf{to} $n_{\mathrm{inner}}$}{
						{Obtain $\bm \phi_i$ by Eq.~(\ref{eqn:FTRML-cls})}\Comment*[r]{inner update}
					}
					{$\bm \phi_{i+1} \gets \bm \phi_i - \alpha \nabla_{\bm \phi_i} \mc L(\bm \phi_i, \mc M^{gm})$}\;
					{{\bf BCL-Dual}: $\bm \theta_t \leftarrow \bm \theta_t + \beta (\bm \phi_{i+1} - \bm \theta_t)$} \Comment*[r]{outer update}
					{{\bf BCL-Single}: $\bm \theta_t \leftarrow \bm \theta_t + \beta (\bm \phi_{1} - \bm \theta_t)$}\Comment*[r]{outer update}
				}
				{$\mc M^{er} \leftarrow $ update $\mc M^{er}$ with $(\vx,y) \in \mc B_j$}\;
			}
			{$\mc M^{er} \leftarrow \mc M^{er} \cup \{f_{\bm \theta_t}(\vx)\}$, $\forall \vx \in \mc M \cap \Dtrain{t}$ }\;
		}
		\Return $\mc \theta_T$
		\caption{Bilevel Continual Learning algorithms.}
		\label{alg:mkd}
	\end{algorithm}
	\vskip -0.2in
\end{figure}

We now present our Bilevel Continual Learning (BCL) algorithm that can simultaneously alleviate catastrophic forgetting and facilitate knowledge transfer.

{\bf Bilevel Continual Learning.}
Given tasks' data arrive in a stream, when receiving a batch of data $\mc B^t_n$ of the task $\mc T_t$, BCL initializes a fast weight $\bm \phi$ from the main model $\bm \theta$ to learn the current data by optimizing Eq.~\ref{eqn:FTRML-cls} and then used to update the main model $\bm \theta_t$ by Eq.(~\ref{eqn:first-order}). The trained fast-weight is discarded before moving to the next batch.
We update both memory units so that the total amount of data stored is within the budget. Any existing memory management strategies \citep{riemer2018learning,lopez2017gradient,rebuffi2017icarl} can be implemented in this step.
Throughout the rest of this paper, we will use {\bf BCL-Dual} to refer to this algorithm that uses two memory units.

{\bf Single Memory Variant of BCL (BCL-Single).}
For comparison, we also develop BCL-Single, a variant of BCL that does not use the generalization memory. As a result, BCL-Single can use only one SGD update, which makes it suitable for a strict online learning scenario where exactly one update per sample is allowed. BCL-Single works by optimizing the inner problem in Eq.~\ref{eqn:FTRML-cls} with exactly one SGD step to obtain the fast weight $\bm \phi_1$. Then, this fast weight is used to update the main model $\bm \theta$ by the same update rule as BCL:
\begin{align}
\bm \theta_t = \bm \theta_t + \beta (\bm \phi_{1} - \bm \theta_t).
\end{align}
Notably, even with one SGD step, BCL-Single differs from normal joint training using experience replay because of the regularizer in Eq.~\ref{eqn:FTRML-cls}.
Alg.~\ref{alg:mkd} gives details of our proposed BCL algorithms.

\vspace*{-0.1in}
\section{Related Work}
\label{sec:related}
{\bf Continual learning}, or lifelong learning, \citep{mccloskey1989catastrophic,ring1997child,thrun1995lifelong} has been extensively studied in literature. Prior works can be broadly categorized into three main categories: (1) regularization, (2) episodic memory, and (3) dynamic architecture.

{\bf (1) Regularization} approaches penalize the changes of influential parameters to previous tasks when learning a new task.
The parameter importance can be estimated by the Fisher information \citep{kirkpatrick2017overcoming} or as the contribution of that parameter to the change of the loss \citep{zenke2017continual} or output \citep{aljundi2017memory}.
However, such approaches find a good solution for all tasks and does not aim at improving the model's generalization.

{\bf (2) Episodic memory} based approaches store a small amount of data from previous tasks and interleaving with data from the current task.
Old data can be used as a constraint to optimize the model \citep{chaudhry2018riemannian,lopez2017gradient,chaudhry2019agem}, representation learning \citep{rebuffi2017icarl}, or just perform joint training with current data \citep{chaudhry2019tiny,riemer2018learning}. Although experience replay and its variants  \citep{castro2018end,belouadah2019il2m,hou2019learning,wu2019large} have achieved promising results, they do not consider the generalization performance of the model, which we focus in this study.

{\bf (3) Dynamic architecture} approaches address catastrophic forgetting by having a separate network for each task and can grow its structure over time \citep{rusu2016progressive,yoon2018lifelong,fernando2017pathnet,li2019learn}. Methods in this category usually do not suffer from catastrophic forgetting because the sub-network of each task is typically frozen. However, they suffer from the unbounded growth of network size, which may not be suitable for some applications.

{\bf Bilevel Optimization} \citep{colson2007overview} refers to a general optimization framework whose constraints involve another optimization problems. Bilevel optimization has been successfully applied in machine learning applications \citep{franceschi2018bilevel,jenni2018deep,liu2018darts} by directly modeling the model's generalization on one set of data (the outer problem) using the knowledge from another set (the inner problem). Our works extend this line of works to the continual learning setting with a dual memory management strategy and a bilevel objective that can prevent catastrophic forgetting as well as facilitate knowledge transfer simultaneously.

\vspace*{-0.1in}
\section{Experiments}
\label{sec:exp}
\subsection{Benchmarks and Baselines}
\vspace*{-0.1in}
We consider four benchmarks in the literature. {\bf Permuted MNISTS} \citep{lopez2017gradient}: each task is a random permutation of the original MNIST.
Here, we generate 23 tasks, each of which has the same amount of training and testing images as the original MNIST data. {\bf Split CIFAR100} \citep{lopez2017gradient} is constructed by splitting the CIFAR100 \citep{krizhevsky2009learning} dataset into 20 tasks, each of which contains five different classes sampled without replacement from the total of 100 classes. Similarly, {\bf Split CUB} and {\bf Split miniImagenet} are constructed by splitting the CUB \citep{WahCUB_200_2011} bird dataset and miniImagenet \citep{vinyals2016matching} dataset into a sequence of 20 tasks, respectively.
In the data pre-processing step, we normalize the images and no other data augmentation is used in all of our experiments.

Throughout the experiments, we compare our {\bf BCL-Dual} and {\bf BCL-Single} with a suite of classic and state-of-the-art methods in the literature. Particularly, we consider the following continual learning methods: {\bf LwF} \citep{li2017learning}, {\bf EWC} \citep{kirkpatrick2017overcoming}, {\bf ICARL} \citep{rebuffi2017icarl}, {\bf GEM} \citep{lopez2017gradient}, {\bf KDR} \citep{hou2018lifelong}, {\bf ER} \citep{chaudhry2019tiny}, {\bf MER} \citep{riemer2018learning}, {\bf FTML} \citep{finn2019online}. We also consider a naive {\bf Finetune} model that trains continuously without any regularization, and an {\bf Offline model} trained on all data of all tasks over three epochs. The Offline model can be viewed as an upper bound of continual learning methods, although it violates the continual learning setting.

\subsection{Implementation Details}
We follow the implementation details proposed in \citep{chaudhry2019agem} in all of our experiments.
Particularly, we use the first three tasks to cross-validate the hyper-parameters of all models and perform continual learning on the remaining tasks.
Moreover, learning is a ``single pass through data'', which means the model only receives each training data once.
We use a small version of Reset18 \citep{he2016deep} (with three times less filter per layer) for Split CIFAR and Split miniImagenet, a pretrained full Resnet18 for Split CUB and a MLP with three hidden layers of 128 neurons for Permuted MNIST.
All methods are optimized by SGD with the mini batch size as 10.
By default, we report BCL-Single with one gradient update so that this version uses the same number of gradient updates with the baselines.
Since BCL-Dual and FTML always require at least 2 gradient updates, we cross-validate the number of gradient updates with the rest of the hyper-parameters using the cross-validation tasks. Following \citep{lopez2017gradient}, we use a single classifier for Permuted MNIST and a separate classifier for each task in the remaining benchmarks.

For GEM, ER, BCL-Single, and BCL-Dual, we use a Ring buffer as the episodic memory's data structure \citep{lopez2017gradient}.
MER and ICARL use reservoir sampling and mean-of-exemplar strategy to maintain their episodic memories as suggested in the original papers.
The total memory size for each task is 256, 65, 50, and 65 in Permuted MNIST, Split CIFAR, Split CUB, and Split miniImagenet respectively.
For BCL-Dual, we use 20\% of the total memory size for the generalization memory in all experiments.
For each benchmark, we run the experiments five times and report the average accuracy (ACC), forgetting measure (FM), and learning accuracy (LA). The details and formulations are provided in Appendix~\ref{sec:protocol}.
\subsection{Results on Standard Benchmarks}
\begin{table}[t!]
	\centering
	\caption{Evaluation metrics of considered methods on four continual learning benchmarks, * denotes the method uses more than {\bf one} gradient update. All methods use the same small Resnet18 backbone and 65 memory slots per task. KDR ran out of memory on Split CUB}
	\label{tab:main-results}
	\setlength\tabcolsep{3.8pt}
	\begin{tabular}{lcccccc}
		\toprule
		\multirow{2}{*}{Method} & \multicolumn{3}{c}{Permutation MNIST}         & \multicolumn{3}{c}{Split CIFAR}            \\
		\cmidrule{2-7}
		& ACC          & FM          & LA           & ACC          & FM           & LA           \\
		\midrule
		Finetune & 32.87$\pm$1.84 & 65.97$\pm$1.93& 95.55$\pm$0.08 & 33.52$\pm$3.13 & 33.88$\pm$2.78 & 65.15$\pm$1.18 \\
		LwF & 39.74$\pm$1.07 & 58.54$\pm$1.14 & 95.36$\pm$0.05 &52.03$\pm$4.11 & 19.34$\pm$4.81 & 64.92$\pm$3.52\\
		EWC    & 56.21$\pm$1.00 & 41.30$\pm$1.07& 95.44$\pm$0.19 & 39.46$\pm$3.75 & 24.69$\pm$3.84 & 64.54$\pm$1.20 \\
		ICARL   & N/A & N/A & N/A & 48.43$\pm$1.73 & 19.63$\pm$1.56 & 66.81$\pm$0.83 \\
		GEM     & 89.51$\pm$0.13 & 6.68$\pm$0.23 & 95.68$\pm$0.10  & 61.36$\pm$0.96 & 7.92$\pm$1.33  & 68.04$\pm$1.42 \\
		KDR & 92.19$\pm$0.19 & 3.92$\pm$0.18 & {95.90$\pm$0.13} & 63.16$\pm$1.02 & 5.17$\pm$1.23 & 66.76$\pm$1.57 \\
		ER     & {88.55$\pm$0.10} & {7.62$\pm$0.11}  & {95.79$\pm$0.03} & 62.23$\pm$0.81 & 7.66$\pm$1.23 & 70.83$\pm$1.35 \\
		MER      & 90.56$\pm$0.12   & 5.80$\pm$0.08    & {95.87$\pm$0.05} &64.36$\pm$0.36&8.06$\pm$0.23& {\bf 71.56$\pm$0.48}\\
		FTML*  & 85.78$\pm$4.34 & 5.84$\pm$1.97 & 91.11$\pm$6.34 & 58.78$\pm$0.93 & 12.48$\pm$0.99 & 70.05$\pm$0.53 \\
		BCL-Single  & {92.17$\pm$0.04} & {\bf 3.01$\pm$0.05} & {95.03$\pm$0.04} & {65.76$\pm$0.93} & {3.61$\pm$1.14}  & {67.05$\pm$1.04} \\
		BCL-Dual*    & {\bf 92.77$\pm$0.10} & {3.59$\pm$0.09} & {\bf 96.18$\pm$0.06} & {\bf67.75$\pm$0.84} & {\bf 2.83$\pm$0.62}  & {69.70$\pm$1.65} \\
		\midrule
		
		Offline & \text95.62$\pm$ 0.07 & - & - & 74.11$\pm$0.66 &- & - \\
		\midrule \midrule
		\multirow{2}{*}{Method} & \multicolumn{3}{c}{Split CUB}             & \multicolumn{3}{c}{Split miniImagenet}     \\
		\cmidrule{2-7}
		& ACC          & FM          & LA           & ACC          & FM           & LA           \\
		\midrule
		Finetune & 64.20$\pm$2.87 & 10.50$\pm$3.68& 71.72$\pm$2.51 & 31.51$\pm$2.00 & 26.00$\pm$2.12 & 55.83$\pm$1.42 \\
		LwF  &65.68$\pm$4.38 & 5.14$\pm$4.76 & 68.21$\pm$3.66 & 43.72$\pm$2.66 & 14.24$\pm$7.40 & 53.03$\pm$3.84 \\
		EWC    & 68.71$\pm$0.88 & 7.62$\pm$0.88 & 74.94$\pm$0.93 & 32.52$\pm$0.53 & 25.74$\pm$2.78 & 56.39$\pm$2.45 \\
		ICARL  & 68.13$\pm$0.87 & 6.95$\pm$1.04 & 74.15$\pm$0.81 & 45.77$\pm$0.62 & 16.29$\pm$0.43 & 61.09$\pm$0.66 \\
		GEM    & 79.72$\pm$2.68 & 3.23$\pm$0.95 & 74.56$\pm$1.19 & 55.30$\pm$1.93 & 5.65$\pm$1.70  & 57.89$\pm$1.44 \\
		KDR  & OOM & OOM & OOM & 58.52$\pm$1.38 & {\bf 4.14$\pm$1.25} & 59.89$\pm$1.73 \\
		ER      & 79.62$\pm$2.68 & 3.55$\pm$2.39 & 75.73$\pm$1.45 & 55.99$\pm$2.53 & 8.27$\pm$2.34  & 60.41$\pm$0.90 \\
		MER    & 82.26$\pm$0.55 & 1.61$\pm$0.31 & 77.96$\pm$0.13 & 58.41$\pm$0.99& 7.97$\pm$1.28 & 65.66$\pm$0.79\\
		FTML* & 78.55$\pm$0.68 & 4.86$\pm$0.40 & {82.02$\pm$0.49} & 51.29$\pm$1.31 & 16.08$\pm$1.59 & {66.42$\pm$0.52} \\
		BCL-Single     & {82.34$\pm$0.48} & {\bf 0.96$\pm$0.71} & {76.76$\pm$1.39} & {59.25$\pm$3.04} & {6.61$\pm$2.85}  & {59.11$\pm$1.50} \\
		BCL-Dual* & {\bf 84.06$\pm$0.40} & {2.80$\pm$0.41} & {\bf 83.85$\pm$0.77} & {\bf63.24$\pm$1.25} & {\bf 4.48$\pm$0.57}  & {\bf 67.15$\pm$0.74} \\
		\midrule
		Offline & 86.56$\pm$2.55 & - & - &71.15$\pm$2.95& - & - \\ \bottomrule
	\end{tabular}
	\vskip -0.2in
\end{table}

Table~\ref{tab:main-results} shows the evaluation metrics for all datasets and methods considered.
Across all baseline methods, FM values are high, indicating that catastrophic forgetting is prominent, greatly reduce their overall performances.
Since FTML requires all data of previous tasks, it does not perform well in continual learning and suffers from catastrophic forgetting.
Although MER can achieve competitive transferring ability with high LA values, it does not balance between knowledge transfer and retaining old knowledge, which leads to lower overall performances compared to our methods.
Our BCL-Single, even with one gradient step, can outperform all baseline considered in terms of overall ACC, FM.
With a dual memory design, BCL-Dual further improves its performance, achieving state-of-the-art results.
Notably, BCL-Dual outperforms other considered methods, even BCL-Single on Split miniImagenet by a large margin. This result show that the dual memory management strategy and bilevel training provide great benefit for continual learning.
The results show that our BCL-Dual and BCL-Single achieves much better performance than the competitors across all benchmarks, confirming our discussion earlier.

\subsection{Adaptation at test time} 

\begin{table}[t]
	\centering
	\caption{Performance of BCL and FTML on Split CIFAR with and without adaptation at test time, all methods use $n=3$ total gradient updates per sample and 256 memory slots per task}
	\label{tab:adapt}
	\begin{tabular}{lP{2cm}P{2cm}P{2cm}}
		\toprule
		\multirow{2}{*}{Method} & \multicolumn{3}{c}{Split CIFAR} \\ \cmidrule{2-4}
		& ACC       & FM       & LA       \\ \midrule
		BCL-Dual                    & { 71.76$\pm$1.77} & { 2.84$\pm$1.76}   & 72.18$\pm$0.31          \\
		BCL-Dual + Adapt            & {\bf 72.06$\pm$0.50} & {\bf 2.08$\pm$0.58}   & {\bf 72.60$\pm$0.42}         \\
		FTML                   & 61.24$\pm$0.65 & 7.23$\pm$0.85   & 67.80$\pm$0.34        \\
		FTML + Adapt           & 63.76$\pm$0.62 & 6.01$\pm$0.72  & 69.24$\pm$0.63          \\ \bottomrule
	\end{tabular}
	\vskip -0.1in
\end{table}
While FTML requires retraining on memory data at test time, we argue that this is an undesired property of a practical continual learning algorithm.
Therefore, we explicitly avoid such need in our methods. However, it is still interesting to explore the benefit of adaptation in continual learning. In this experiment, we sidestep the conventional continual learning setting by allowing the model to finetune on the memory of a task before evaluation on that task. 

We compare our BCL-Dual with FTML on the Split CIFAR benchmark with 256 memory slots per task and report their results in Table~\ref{tab:adapt}. We choose a larger memory size because we did not observe significant improvements with only 65 memory slots. While FTML's performance improved when adaptation is allowed, it still suffers from catastrophic forgetting indicated by high FM values. Moreover, even with adaptation, FTML still performs worse than BCL-Dual. On the other hand, BCL-Dual performs consistently with and without adaptation at test time. This shows that our BCL-Dual is robust to adaptation, making it a suitable method for practical continual learning.

\subsection{Ablation Studies}
\begin{table*}[t]
	\centering 
	\caption{Contribution of each component in BCL-Dual on  Split CIFAR benchmark, Reg: regularization in Eq.~\ref{eqn:FTRML-cls}, 2SGD: 3 inner update steps, DM: dual memory management strategy}
	\label{tab:ablation}
	\vskip 0.1in
	\setlength\tabcolsep{3.8pt}
	\begin{tabular}{lcccccc}
		\toprule
		\multirow{2}{*}{Method} & \multirow{2}{*}{Reg} & \multirow{2}{*}{2SGD} & \multirow{2}{*}{DM} & \multicolumn{3}{c}{Split CIFAR}                      \\ \cmidrule{5-7}
		&                      &                       &                              & ACC                  & FM            & LA            \\ \midrule
		BCL-Dual                     & $\checkmark$         & $\checkmark$          & $\checkmark$                 & {\bf 67.75$\pm$0.84} & {\bf 2.83$\pm$0.62}  & {69.70$\pm$1.65} \\
		&                      & $\checkmark$          & $\checkmark$                 & 63.71$\pm$0.39         & 6.85$\pm$0.76   & 69.93$\pm$0.35  \\
		& $\checkmark$         &                       &                              & {65.76$\pm$0.93} & {3.61$\pm$1.14}  & {67.05$\pm$1.04}  \\
		& $\checkmark$         & $\checkmark$          &                              & 66.74$\pm$0.60         & 3.94$\pm$0.51   & 70.21$\pm$0.50  \\ \midrule
		ER                        &                      &                       &                              & 62.23$\pm$0.81         & 7.66$\pm$1.23   & 70.83$\pm$1.35  \\ \bottomrule
	\end{tabular}
	\vskip -0.1in
\end{table*}

We conduct various ablation studies to further understand BCL-Dual. Mainly, we are interested in examining the contribution of each component in BCL-Dual and its performance with different numbers of training samples per task.

First, we study how each BCL's component contributes to its overall performance. Particularly, we want to study the contribution of the regularization in Eq.~\ref{eqn:FTRML-cls}, the benefit of doing several inner updates in Eq.~\ref{eqn:maml-inner} as well as the dual memory management strategy for the outer update in Eq.~\ref{eqn:maml-outer}.
We consider the Split CIFAR benchmark with 65 memory slots per task for this experiment and report the results in Table.~\ref{tab:ablation}. Notably, BCL-Dual with only the regularization (Reg) is equivalent to BCL-Single. When BCL-Single is trained with more inner updates (Reg+2SGD), it only differs from BCCL-Dual in the dual memory. The results show that the dual memory management can offer 1\% ACC improvements to BCL. Moreover, the regularization helps address the bias caused by small episodic memory sizes in BCL and improves the overall performance. 

\begin{figure*}[t!]
	\centering
	
	\begin{subfigure}[t]{0.44\textwidth}
		\includegraphics[width=\textwidth]{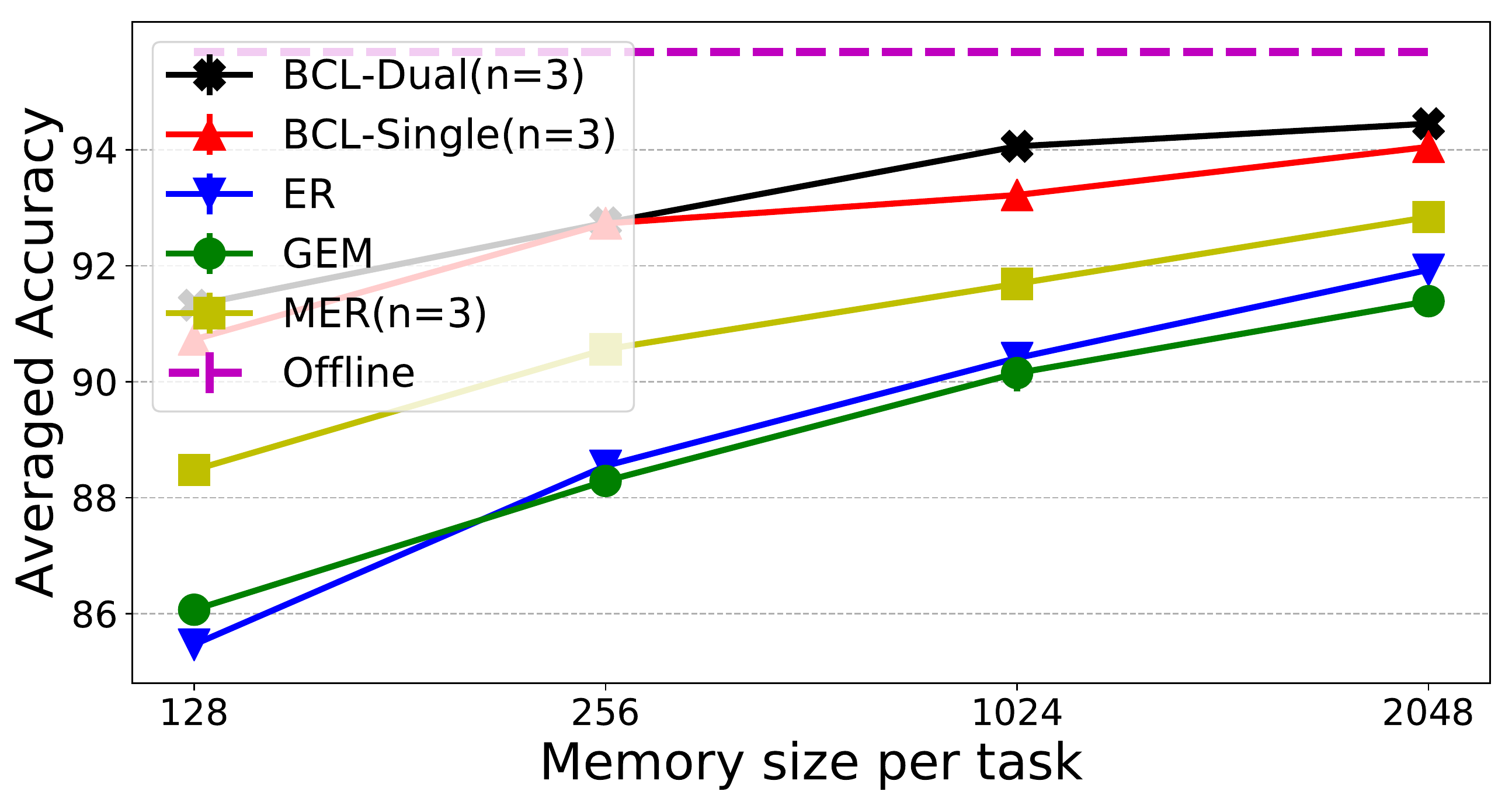}
		\caption{Permuted MNIST}
	\end{subfigure}
	~
	\begin{subfigure}[t]{0.44\textwidth}
		\includegraphics[width=\textwidth]{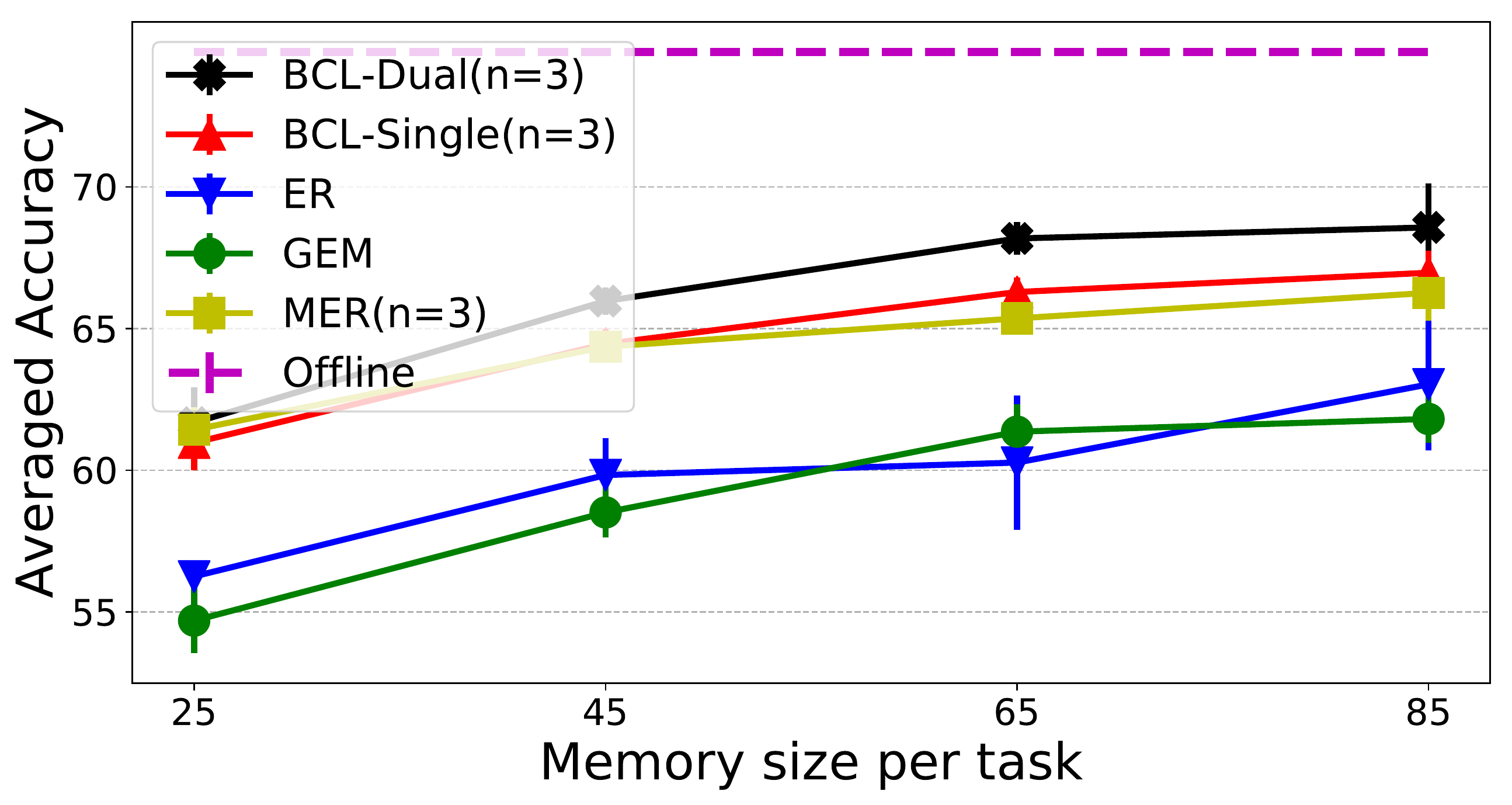}
		\caption{Split CIFAR}
	\end{subfigure}
	\\
	\begin{subfigure}[t]{0.44\textwidth}
		\includegraphics[width=\textwidth]{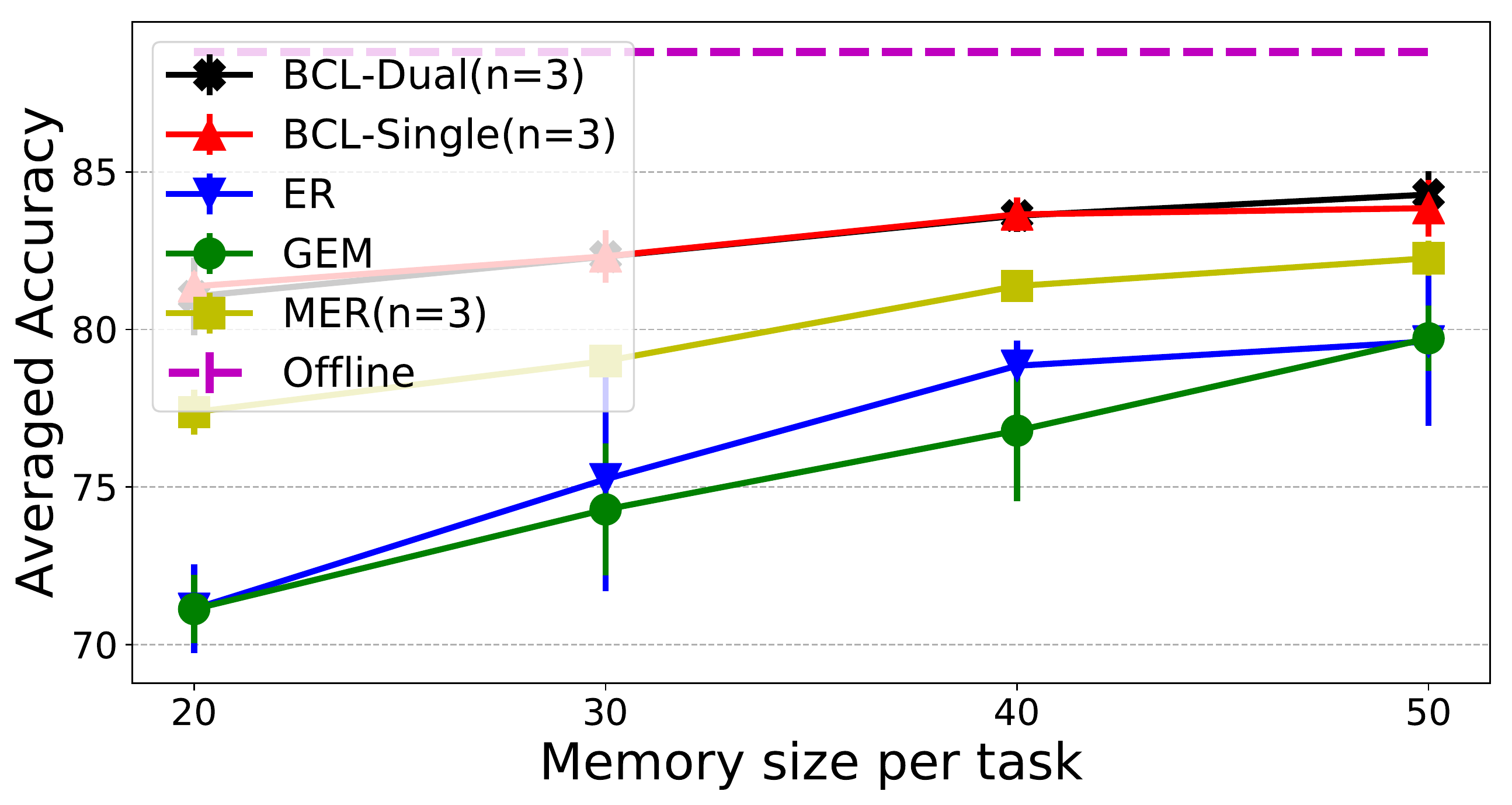}
		\caption{Split CUB}
	\end{subfigure}
	~
	\begin{subfigure}[t]{0.44\textwidth}
		\includegraphics[width=\textwidth]{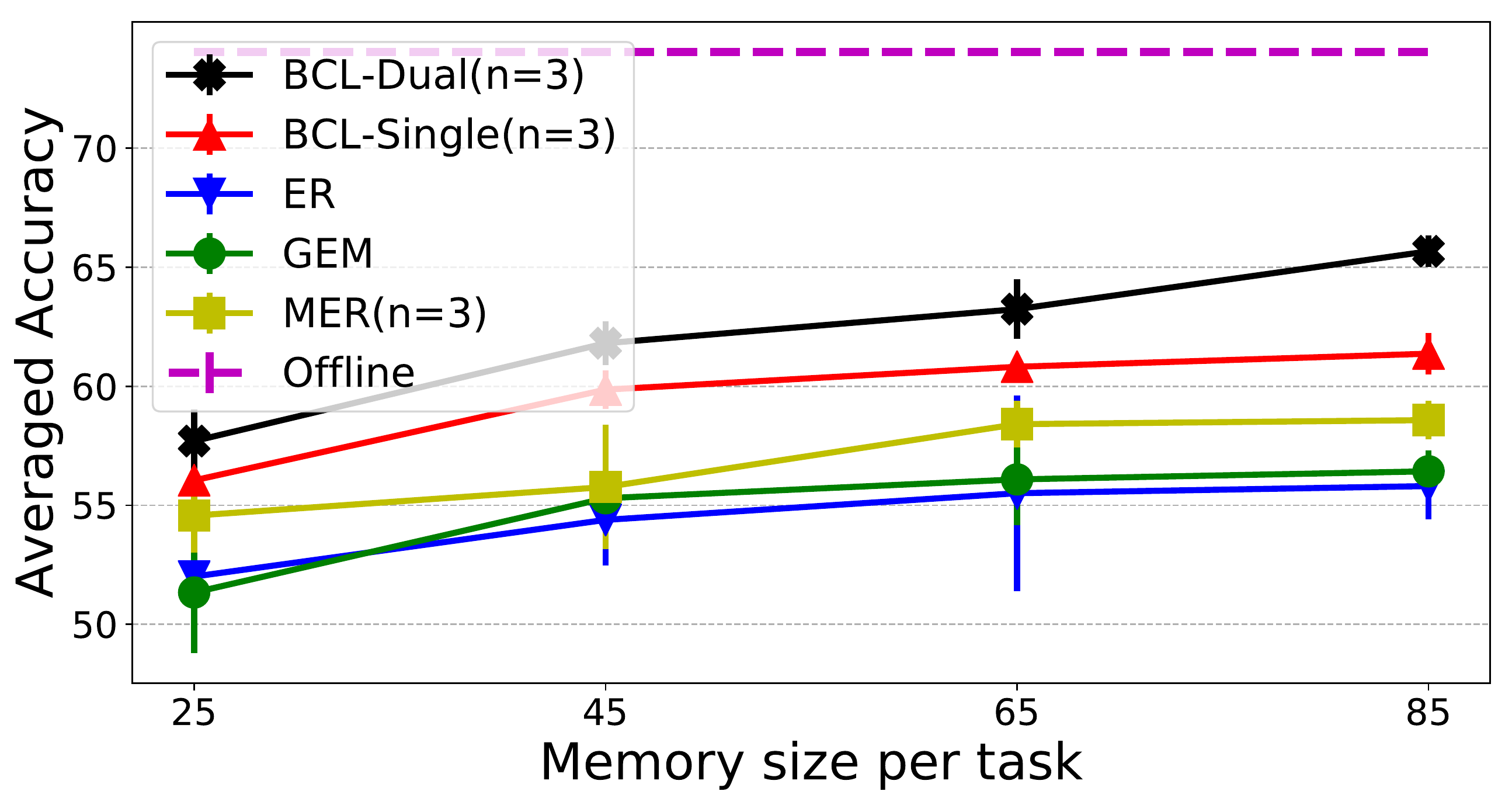}
		\caption{Split miniImagenet}
	\end{subfigure}
	\caption{Comparison of the effect of the episodic memory size in four continual learning benchmarks. ($n=3$) denotes the method is optimized with three SGD steps.  Error bars on Permuted MNIST are not visible due to small standard deviation values, i.e. $<0.05$. Best viewed in colors}
	\label{fig:memsz}
\end{figure*}

Finally, we study the effect of the episodic memory size on different continual learning methods.
We evaluate four competing methods GEM, ER, MER, BCL-Single and BCL-Dual on four benchmarks: Permuted MNIST, Split CIFAR, Split CUB, and Split miniImagenet. Fig.~\ref{fig:memsz} plots the average accuracy curve of each benchmark against the episodic memory size. We observe that, in all cases, the performance of all four methods generally improves when the episodic memory size increases, which is easy to understand as a larger memory will lead to a better representative of the original training data. For BCL-Single, we use two SGD updates, which means it only differs from BCL-Single in the dual memory management. We observe that the dual memory management strategy in BCL consistently offers 0.5\% to 2\% ACC improvements over BCL-Single across all benchmarks. The results show that BCL-Dual's bilevel learning and dual memory can outperform existing methods across all benchmarks and different memory sizes.

\section{Conclusion}
\label{sec:conclusion}

In this paper, we have investigated the potentials and limitations of existing continual learning methods. We have shown that while they can achieve reasonably competitive performance on some benchmarks, catastrophic forgetting remains a challenge, and transferring knowledge among tasks is not fully addressed, especially in the online continual learning setting. These limitations motivated us to propose a new framework for continual learning through a novel bilevel optimization approach and a dual memory management strategy. Based on a first-order approximation, our BCL algorithms strike a great balance between preventing catastrophic forgetting and facilitating learning to future tasks. Through extensive experiments on popular continual learning benchmarks, our methods consistently achieved state-of-the-art results on different memory sizes.

\bibliography{iclr2020_conference}

\begin{thebibliography}{39}
\providecommand{\natexlab}[1]{#1}
\providecommand{\url}[1]{\texttt{#1}}
\expandafter\ifx\csname urlstyle\endcsname\relax
  \providecommand{\doi}[1]{doi: #1}\else
  \providecommand{\doi}{doi: \begingroup \urlstyle{rm}\Url}\fi

\bibitem[Aljundi et~al.(2018)Aljundi, Babiloni, Elhoseiny, Rohrbach, and
  Tuytelaars]{aljundi2017memory}
Rahaf Aljundi, Francesca Babiloni, Mohamed Elhoseiny, Marcus Rohrbach, and
  Tinne Tuytelaars.
\newblock Memory aware synapses: Learning what (not) to forget.
\newblock In \emph{Proceedings of the European Conference on Computer Vision
  (ECCV)}, pp.\  139--154, 2018.

\bibitem[Belouadah \& Popescu(2019)Belouadah and Popescu]{belouadah2019il2m}
Eden Belouadah and Adrian Popescu.
\newblock Il2m: Class incremental learning with dual memory.
\newblock In \emph{Proceedings of the IEEE International Conference on Computer
  Vision}, pp.\  583--592, 2019.

\bibitem[Castro et~al.(2018)Castro, Mar{\'\i}n-Jim{\'e}nez, Guil, Schmid, and
  Alahari]{castro2018end}
Francisco~M Castro, Manuel~J Mar{\'\i}n-Jim{\'e}nez, Nicol{\'a}s Guil, Cordelia
  Schmid, and Karteek Alahari.
\newblock End-to-end incremental learning.
\newblock In \emph{Proceedings of the European Conference on Computer Vision
  (ECCV)}, pp.\  233--248, 2018.

\bibitem[Chaudhry et~al.(2018)Chaudhry, Dokania, Ajanthan, and
  Torr]{chaudhry2018riemannian}
Arslan Chaudhry, Puneet~K Dokania, Thalaiyasingam Ajanthan, and Philip~HS Torr.
\newblock Riemannian walk for incremental learning: Understanding forgetting
  and intransigence.
\newblock In \emph{Proceedings of the European Conference on Computer Vision
  (ECCV)}, pp.\  532--547, 2018.

\bibitem[Chaudhry et~al.(2019{\natexlab{a}})Chaudhry, Ranzato, Rohrbach, and
  Elhoseiny]{chaudhry2019agem}
Arslan Chaudhry, Marc'Aurelio Ranzato, Marcus Rohrbach, and Mohamed Elhoseiny.
\newblock Efficient lifelong learning with a-gem.
\newblock \emph{International Conference on Learning Representations (ICLR)},
  2019{\natexlab{a}}.

\bibitem[Chaudhry et~al.(2019{\natexlab{b}})Chaudhry, Rohrbach, Elhoseiny,
  Ajanthan, Dokania, Torr, and Ranzato]{chaudhry2019tiny}
Arslan Chaudhry, Marcus Rohrbach, Mohamed Elhoseiny, Thalaiyasingam Ajanthan,
  Puneet~K Dokania, Philip~HS Torr, and Marc'Aurelio Ranzato.
\newblock On tiny episodic memories in continual learning.
\newblock \emph{arXiv preprint arXiv:1902.10486}, 2019{\natexlab{b}}.

\bibitem[Colson et~al.(2007)Colson, Marcotte, and Savard]{colson2007overview}
Beno{\^\i}t Colson, Patrice Marcotte, and Gilles Savard.
\newblock An overview of bilevel optimization.
\newblock \emph{Annals of operations research}, 153\penalty0 (1):\penalty0
  235--256, 2007.

\bibitem[Domke(2012)]{domke2012generic}
Justin Domke.
\newblock Generic methods for optimization-based modeling.
\newblock In \emph{Artificial Intelligence and Statistics}, pp.\  318--326,
  2012.

\bibitem[Fernando et~al.(2017)Fernando, Banarse, Blundell, Zwols, Ha, Rusu,
  Pritzel, and Wierstra]{fernando2017pathnet}
Chrisantha Fernando, Dylan Banarse, Charles Blundell, Yori Zwols, David Ha,
  Andrei~A Rusu, Alexander Pritzel, and Daan Wierstra.
\newblock Pathnet: Evolution channels gradient descent in super neural
  networks.
\newblock \emph{arXiv preprint arXiv:1701.08734}, 2017.

\bibitem[Finn et~al.(2019)Finn, Rajeswaran, Kakade, and Levine]{finn2019online}
Chelsea Finn, Aravind Rajeswaran, Sham Kakade, and Sergey Levine.
\newblock Online meta-learning.
\newblock In \emph{Proceedings of the 36th International Conference on Machine
  Learning-Volume 97}, pp.\  1920--1930. JMLR. org, 2019.

\bibitem[Franceschi et~al.(2017)Franceschi, Donini, Frasconi, and
  Pontil]{franceschi2017forward}
Luca Franceschi, Michele Donini, Paolo Frasconi, and Massimiliano Pontil.
\newblock Forward and reverse gradient-based hyperparameter optimization.
\newblock In \emph{Proceedings of the 34th International Conference on Machine
  Learning-Volume 70}, pp.\  1165--1173. JMLR. org, 2017.

\bibitem[Franceschi et~al.(2018)Franceschi, Frasconi, Salzo, Grazzi, and
  Pontil]{franceschi2018bilevel}
Luca Franceschi, Paolo Frasconi, Saverio Salzo, Riccardo Grazzi, and
  Massimiliano Pontil.
\newblock Bilevel programming for hyperparameter optimization and
  meta-learning.
\newblock In \emph{International Conference on Machine Learning}, pp.\
  1568--1577, 2018.

\bibitem[French(1999)]{french1999catastrophic}
Robert~M French.
\newblock Catastrophic forgetting in connectionist networks.
\newblock \emph{Trends in cognitive sciences}, 3\penalty0 (4):\penalty0
  128--135, 1999.

\bibitem[He et~al.(2016)He, Zhang, Ren, and Sun]{he2016deep}
Kaiming He, Xiangyu Zhang, Shaoqing Ren, and Jian Sun.
\newblock Deep residual learning for image recognition.
\newblock In \emph{Proceedings of the IEEE conference on computer vision and
  pattern recognition}, pp.\  770--778, 2016.

\bibitem[Hinton et~al.(2015)Hinton, Vinyals, and Dean]{hinton2015distilling}
Geoffrey Hinton, Oriol Vinyals, and Jeff Dean.
\newblock Distilling the knowledge in a neural network.
\newblock In \emph{NIPS Deep Learning and Representation Learning Workshop},
  2015.
\newblock URL \url{http://arxiv.org/abs/1503.02531}.

\bibitem[Hou et~al.(2018)Hou, Pan, Change~Loy, Wang, and Lin]{hou2018lifelong}
Saihui Hou, Xinyu Pan, Chen Change~Loy, Zilei Wang, and Dahua Lin.
\newblock Lifelong learning via progressive distillation and retrospection.
\newblock In \emph{Proceedings of the European Conference on Computer Vision
  (ECCV)}, pp.\  437--452, 2018.

\bibitem[Hou et~al.(2019)Hou, Pan, Loy, Wang, and Lin]{hou2019learning}
Saihui Hou, Xinyu Pan, Chen~Change Loy, Zilei Wang, and Dahua Lin.
\newblock Learning a unified classifier incrementally via rebalancing.
\newblock In \emph{Proceedings of the IEEE Conference on Computer Vision and
  Pattern Recognition}, pp.\  831--839, 2019.

\bibitem[Jenni \& Favaro(2018)Jenni and Favaro]{jenni2018deep}
Simon Jenni and Paolo Favaro.
\newblock Deep bilevel learning.
\newblock In \emph{Proceedings of the European Conference on Computer Vision
  (ECCV)}, pp.\  618--633, 2018.

\bibitem[Kirkpatrick et~al.(2017)Kirkpatrick, Pascanu, Rabinowitz, Veness,
  Desjardins, Rusu, Milan, Quan, Ramalho, Grabska-Barwinska,
  et~al.]{kirkpatrick2017overcoming}
James Kirkpatrick, Razvan Pascanu, Neil Rabinowitz, Joel Veness, Guillaume
  Desjardins, Andrei~A Rusu, Kieran Milan, John Quan, Tiago Ramalho, Agnieszka
  Grabska-Barwinska, et~al.
\newblock Overcoming catastrophic forgetting in neural networks.
\newblock \emph{Proceedings of the national academy of sciences}, 2017.

\bibitem[Krizhevsky \& Hinton(2009)Krizhevsky and
  Hinton]{krizhevsky2009learning}
Alex Krizhevsky and Geoffrey Hinton.
\newblock Learning multiple layers of features from tiny images.
\newblock Technical report, Citeseer, 2009.

\bibitem[Li et~al.(2019)Li, Zhou, Wu, Socher, and Xiong]{li2019learn}
Xilai Li, Yingbo Zhou, Tianfu Wu, Richard Socher, and Caiming Xiong.
\newblock Learn to grow: A continual structure learning framework for
  overcoming catastrophic forgetting.
\newblock 2019.

\bibitem[Li \& Hoiem(2017)Li and Hoiem]{li2017learning}
Zhizhong Li and Derek Hoiem.
\newblock Learning without forgetting.
\newblock \emph{IEEE Transactions on Pattern Analysis and Machine
  Intelligence}, 2017.

\bibitem[Liu et~al.(2018)Liu, Simonyan, and Yang]{liu2018darts}
Hanxiao Liu, Karen Simonyan, and Yiming Yang.
\newblock Darts: Differentiable architecture search.
\newblock \emph{arXiv preprint arXiv:1806.09055}, 2018.

\bibitem[Lopez-Paz \& Ranzato(2017)Lopez-Paz and Ranzato]{lopez2017gradient}
David Lopez-Paz and Marc'Aurelio Ranzato.
\newblock Gradient episodic memory for continual learning.
\newblock In \emph{Advances in Neural Information Processing Systems}, pp.\
  6467--6476, 2017.

\bibitem[McCloskey \& Cohen(1989)McCloskey and
  Cohen]{mccloskey1989catastrophic}
Michael McCloskey and Neal~J Cohen.
\newblock Catastrophic interference in connectionist networks: The sequential
  learning problem.
\newblock In \emph{Psychology of learning and motivation}, volume~24, pp.\
  109--165. Elsevier, 1989.

\bibitem[Nichol et~al.(2018)Nichol, Achiam, and Schulman]{nichol2018first}
Alex Nichol, Joshua Achiam, and John Schulman.
\newblock On first-order meta-learning algorithms.
\newblock \emph{arXiv preprint arXiv:1803.02999}, 2018.

\bibitem[Parisi et~al.(2019)Parisi, Kemker, Part, Kanan, and
  Wermter]{parisi2019continual}
German~I Parisi, Ronald Kemker, Jose~L Part, Christopher Kanan, and Stefan
  Wermter.
\newblock Continual lifelong learning with neural networks: A review.
\newblock \emph{Neural Networks}, 2019.

\bibitem[Rebuffi et~al.(2017)Rebuffi, Kolesnikov, Sperl, and
  Lampert]{rebuffi2017icarl}
Sylvestre-Alvise Rebuffi, Alexander Kolesnikov, Georg Sperl, and Christoph~H
  Lampert.
\newblock icarl: Incremental classifier and representation learning.
\newblock In \emph{Proceedings of the IEEE Conference on Computer Vision and
  Pattern Recognition}, pp.\  2001--2010, 2017.

\bibitem[Riemer et~al.(2019)Riemer, Cases, Ajemian, Liu, Rish, Tu, and
  Tesauro]{riemer2018learning}
Matthew Riemer, Ignacio Cases, Robert Ajemian, Miao Liu, Irina Rish, Yuhai Tu,
  and Gerald Tesauro.
\newblock Learning to learn without forgetting by maximizing transfer and
  minimizing interference.
\newblock \emph{International Conference on Learning Representations (ICLR)},
  2019.

\bibitem[Ring(1997)]{ring1997child}
Mark~B Ring.
\newblock Child: A first step towards continual learning.
\newblock \emph{Machine Learning}, 28\penalty0 (1):\penalty0 77--104, 1997.

\bibitem[Rusu et~al.(2016)Rusu, Rabinowitz, Desjardins, Soyer, Kirkpatrick,
  Kavukcuoglu, Pascanu, and Hadsell]{rusu2016progressive}
Andrei~A Rusu, Neil~C Rabinowitz, Guillaume Desjardins, Hubert Soyer, James
  Kirkpatrick, Koray Kavukcuoglu, Razvan Pascanu, and Raia Hadsell.
\newblock Progressive neural networks.
\newblock \emph{arXiv preprint arXiv:1606.04671}, 2016.

\bibitem[Sahoo et~al.(2018)Sahoo, Pham, Lu, and Hoi]{sahoo2018online}
Doyen Sahoo, Quang Pham, Jing Lu, and Steven C.~H. Hoi.
\newblock Online deep learning: Learning deep neural networks on the fly.
\newblock In \emph{Proceedings of the Twenty-Seventh International Joint
  Conference on Artificial Intelligence, {IJCAI-18}}, 2018.

\bibitem[Thrun \& Mitchell(1995)Thrun and Mitchell]{thrun1995lifelong}
Sebastian Thrun and Tom~M Mitchell.
\newblock Lifelong robot learning.
\newblock In \emph{The biology and technology of intelligent autonomous
  agents}, pp.\  165--196. Springer, 1995.

\bibitem[Vinyals et~al.(2016)Vinyals, Blundell, Lillicrap, Wierstra,
  et~al.]{vinyals2016matching}
Oriol Vinyals, Charles Blundell, Timothy Lillicrap, Daan Wierstra, et~al.
\newblock Matching networks for one shot learning.
\newblock In \emph{Advances in neural information processing systems}, pp.\
  3630--3638, 2016.

\bibitem[Wah et~al.(2011)Wah, Branson, Welinder, Perona, and
  Belongie]{WahCUB_200_2011}
C.~Wah, S.~Branson, P.~Welinder, P.~Perona, and S.~Belongie.
\newblock {The Caltech-UCSD Birds-200-2011 Dataset}.
\newblock Technical Report CNS-TR-2011-001, California Institute of Technology,
  2011.

\bibitem[Wu et~al.(2019)Wu, Chen, Wang, Ye, Liu, Guo, and Fu]{wu2019large}
Yue Wu, Yinpeng Chen, Lijuan Wang, Yuancheng Ye, Zicheng Liu, Yandong Guo, and
  Yun Fu.
\newblock Large scale incremental learning.
\newblock In \emph{Proceedings of the IEEE Conference on Computer Vision and
  Pattern Recognition}, pp.\  374--382, 2019.

\bibitem[Yoon et~al.(2018)Yoon, Yang, Lee, and Hwang]{yoon2018lifelong}
Jaehong Yoon, Eunho Yang, Jeongtae Lee, and Sung~Ju Hwang.
\newblock Lifelong learning with dynamically expandable networks.
\newblock \emph{International Conference on Learning Representations (ICLR)},
  2018.

\bibitem[Zenke et~al.(2017)Zenke, Poole, and Ganguli]{zenke2017continual}
Friedemann Zenke, Ben Poole, and Surya Ganguli.
\newblock Continual learning through synaptic intelligence.
\newblock In \emph{Proceedings of the 34th International Conference on Machine
  Learning-Volume 70}, pp.\  3987--3995. JMLR. org, 2017.

\bibitem[Zhang et~al.(2019)Zhang, Lucas, Ba, and Hinton]{zhang2019lookahead}
Michael Zhang, James Lucas, Jimmy Ba, and Geoffrey~E Hinton.
\newblock Lookahead optimizer: k steps forward, 1 step back.
\newblock In \emph{Advances in Neural Information Processing Systems}, pp.\
  9593--9604, 2019.

\end{thebibliography}
\bibliographystyle{iclr2020_conference}

\appendix
\section{Continual Learning Protocol and Evaluation Metrics}
\label{sec:protocol}
In this section, we introduce the online continual learning problem and then present the evaluation metrics. 
We use $\mc T_t$ and $\mc D_t$ to indicate the $t$-th task and its training dataset. Particularly, ${\mc D^{tr}_t = \{(\vx^{n}_t, y^{n}_t,t)\}_{n=1}^{N_t}}$ is the training set of task $\mc T_t$ and each sample $(\vx^n_t, y^n_t,t)$ includes an input vector $\vx^{n}_t$, a target vector $y^{n}_t$, and a task identifier $t$.
Similarly, $\mc D^{te}_t$ denotes the testing set of task $\mc T_t$. We denote the generalization memory $\mc M^{gm}$ as a set of data sampled from the training data of all observed tasks so far but not used to directly train the model. The goal of learning is to construct a predictive model $f_{\bm \theta}$ parameterized by $\bm \theta$ such that it can predict the target vector $y \approx f_{\bm \theta}(\vx, t)$ associated to an unseen input $\vx$ from any of the observed tasks.

In continual learning setting, the model  observes a sequence of $T$ tasks \\ $\mc{T} = \{\mc T_1, \ldots,\mc T_t, \ldots, \mc T_T\}$ in a streaming way.
At any given time $t$, only task $\mc T_t$ is presented to the learner, and it has to learn to solve all observed tasks without accessing to the previous tasks' data.
In this work, we follow the protocol proposed in \citep{lopez2017gradient}, where the data within each task also arrive sequentially and the task index is also given as part of the input.
Knowing the task index is equivalent to the ``multi-head'' evaluation \citep{chaudhry2018riemannian} in which only the corresponding classifier to the presented task is evaluated at test time.
The goal of continual learning is to obtain a model that performs well on the current task $\mc T_t$ and all of the previous tasks $\mc T_{<t}$ at any time $t$. We also allow the use of an episodic memory $\mc M$ to store some useful information on previous tasks such as a small amount of data for experience replay.

For a principled evaluation, we adopt three standard metrics in the literature: Average Accuracy (ACC) \citep{lopez2017gradient}, Forgetting Measure (FM) \citep{chaudhry2018riemannian}, and Learning Accuracy (LA) \citep{riemer2018learning}. At any time $t$ in training phase, we denote $a_{i,j}$ as the model's accuracy evaluated on the test set $\mathcal{D}_j^{te}$ of task $\mc T_j$ after it has been trained on the last sample in dataset $\mathcal{D}^{tr}_i$ of task $\mc T_i$. Then, the three metrics are defined as follows:
\begin{itemize}[leftmargin=*]
	\item {\bf Averaged Accuracy:} $\displaystyle \text{ACC} = \frac{1}{T} \sum_{i = 1}^T a_{T,i}.$
	\item{\bf Forgetting Measure: } $\displaystyle \text{FM} = \frac{1}{T-1} \sum_{j=1}^{T-1} \max_{l < T }a_{l,j} - a_{T,j}.$
	\item{\bf Learning Accuracy:} $\displaystyle \text{LA} = \frac{1}{T} \sum_{i=1}^T a_{i,i}.$
\end{itemize}
The presented metrics measure different aspects of continual learning.
The average accuracy shows the model's performance on observed tasks at the end of the training.
Forgetting metric measures the model's ability to retain prior knowledge when it learns new information. 
Finally, learning accuracy evaluates the model's ability to use its old knowledge to improve the learning of future tasks, reflecting its ability to transfer. In general, the overall performances of any two methods are compared via ACC. If we want to explicitly look at how much the model forgot or how well it can transfer knowledge, then FM or LA will be used.

\section{Class-Incremental Learning Experiment}
\begin{table}[h]
	\centering
	\caption{Accuracy (ACC) and Forgetting Measure (FM) of considered methods on the CIFAR-10 benchmark. A Small Resnet18 backbone is used in all methods. M denotes the amount of data stored per class, $^{\dagger}$ denotes the result is collected from \cite{aljundi2019online}, -RAND suffix denotes random sampling, -MIR suffix denotes MIR sampling. Best results of each sampling strategy are highlighted in bold}
	\label{tab:ocil}
	\begin{tabular}{lP{1.5cm}P{1.5cm}P{1.5cm}P{1.5cm}}
		\toprule
		\multirow{2}{*}{Method}          & \multicolumn{2}{c}{M=50}                &  \multicolumn{2}{c}{M=50}         \\ \cmidrule{2-5}
		& ACC        & FM         & ACC        & FM         \\ \midrule
		offline$^{\dagger}$    & 79.2$\pm$0.4 & N/A        & N/A        & N/A        \\
		finetune$^{\dagger}$   & 18.4$\pm$0.3 & 85.4$\pm$0.7 & 18.4$\pm$0.3 & 85.4$\pm$0.7 \\ \midrule
		GEM-RAND$^{\dagger}$        & 17.1$\pm$1.0 & 70.7$\pm$4.5 & 17.5$\pm$1.6 & 71.7$\pm$1.3 \\
		ICARL-RAND$^{\dagger}$      & 33.7$\pm$1.6 & 40.6$\pm$1.1 & 32.4$\pm$2.1 & 40.8$\pm$1.8 \\
		ER-RAND$^{\dagger}$         & 33.1$\pm$1.7 & 35.4$\pm$2.0 & 41.3$\pm$1.9 & {\bf 23.3$\pm$2.9} \\
		BCL-FO-RAND     & {\bf 39.5$\pm$2.2} & {\bf 28.3$\pm$3.6} & {\bf 43.8$\pm$1.2} & {\bf 23.1$\pm$2.7} \\ \midrule
		ER-MIR$^{\dagger}$     & 40.0$\pm$1.1 & 30.2$\pm$2.3 & 47.6$\pm$1.1 & {17.4$\pm$2.1} \\
		BCL-FO-MIR & {\bf 43.6$\pm$1.6} & {\bf 28.0$\pm$2.2} & {\bf 48.2$\pm$0.7}  & {\bf 16.5$\pm$1.6} \\ \bottomrule
	\end{tabular}
\end{table}

To demonstrate this property, we consider the online class-incremental learning protocol \cite{aljundi2019online} in which the task identifier is not given to the model and it has to make predictions on all observed classes so far and consider the state-of-the-art method: {\it Maximally Interfered Retrieval} (MIR). Instead of randomly sample a mini batch of data from the memory at each step, MIR works by selecting the replay data that maximize the model's forgetting by performing a virtual update in each step. Therefore, it directly aims at reducing the model's forgetting measure, which is a challenge in online Class-Incremental learning.

In this experiment, we show that by replacing the random sampling strategy in BCL with MIR sampling, we directly observe improvements under the same setting in the CIFAR-10 benchmark used in \cite{aljundi2019online}.
For a fair comparison, we implement our BCL-FO on their publicly available implementation\footnote{\url{https://github.com/optimass/Maximally_Interfered_Retrieval}}
and reuse all their setting such as data split, training, and evaluation protocols.
Due to time constraints and our results on Split CIFAR100 showed that BCL and BCL-FO achieved similar performances on the CIFAR100 dataset, we only consider BCL-FO in this experiment.

We compare BCL-FO with and without MIR sampling strategies with the baselines in \cite{aljundi2019online} and report the accuracy at the end of learning (ACC), forgetting measure (FM) in Table~\ref{tab:ocil}. We observe that BCL-FO-RAND consistently outperforms other methods with random sampling strategies and even comes close to ER-MIR which uses MIR sampling when 50 memory slots per class are allowed. When we replace random sampling in BCL-FO with MIR sampling, BCl-FO-MIR outperforms all the methods considered in both memory sizes, including ER-MIR. This experiment demonstrates that our BCL framework and existing, orthogonal works are complementary to each other and they can work collectively together to achieve new state-of-the-art results with minimal modification.
\section{Experiment Details}
One challenge of the online continual learnin is that we are not allowed to perform hyper-parameters search on new tasks data because they are not available prior to learning. Therefore, cross-validating the hyper-parameters on the validation data of all tasks may be an optimistic estimation of the model's performance. In practice, this step should be performed prior to actual continual learning, as discussed in \cite{chaudhry2019agem} and followed in our experiments. In this section, we summarize and report the hyper-parameter setting used in all experiments conducted in this work for reproducibility and future research. We follow the same notation used in Algorithm 1 in the main paper. Our implementation is available at \href{https://github.com/phquang/bilevel-continual-learning}{https://github.com/phquang/bilevel-continual-learning}.
\subsection{BCL-Dual and BCL-Single}
\begin{itemize}
	\item Inner learning rate: 0.03 (Permutation MNIST, Split CUB), 0.3 (Split CIFAR), 0.05 (Split miniImagenet)
	\item Outer learning rate $\beta$: $\beta$: 0.3 (Permutation MNIST, Split CUB), 0.1 (Split CIFAR, Split miniImagenet)
	\item Replay batch size: 128 (Permutation MNIST, Split CIFAR, Split CUB, Split miniImagenet)
	\item Validation set size: 20\% of total memory size
	\item Temperature $\tau$: 5 (Permutation MNIST, Split CIFAR, Split CUB, Split miniImagenet)
	\item Regularization $\lambda$: 100 (Permutation MNIST, Split CIFAR, Split CUB, Split miniImagenet)
	\item Inner loops $n_{\text{inner}}$: 2 (Permutation MNIST, Split CIFAR), 3 (Split CUB, Split miniImagenet)
	\item Outer loops $n_{\text{outer}}$: 1 (Permutation MNIST, Split CIFAR, Split miniImagenet), 2 (Split CUB).
\end{itemize}

\end{document}